\pdfoutput=1

\documentclass[11pt]{article}

\usepackage[]{acl}
\usepackage{times}
\usepackage{latexsym}
\usepackage{graphicx,xcolor}  
\usepackage{amsmath} 
\usepackage[whole]{bxcjkjatype}
\usepackage[subrefformat=parens]{subcaption}
\usepackage{here} 
\usepackage{booktabs}
\usepackage[T1]{fontenc}

\usepackage[utf8]{inputenc}

\usepackage{microtype}

\title{Tracing and Manipulating Intermediate Values \\
in Neural Math Problem Solvers}

\author{
  Yuta Matsumoto$^{\,1}$ ~~
  Benjamin Heinzerling$^{\,2,1}$ ~~
  Masashi Yoshikawa$^{\,1}$ ~~
  Kentaro Inui$^{\,1,2}$ \\
  ${}^{1}$Tohoku University / Japan ~~ ${}^{2}$RIKEN / Japan \\
  \texttt{yuta.matsumoto.q8@dc.tohoku.ac.jp}, \quad
  \texttt{benjamin.heinzerling@riken.jp}, \\
  \texttt{yoshikawa@tohoku.ac.jp}, \quad
  \texttt{inui@ecei.tohoku.ac.jp},
}

\begin{document}
\maketitle
\begin{abstract}
How language models process complex input that requires multiple steps of inference is not well understood.
Previous research has shown that information about intermediate values of these inputs can be extracted from the activations of the models, but it is unclear where that information is encoded and whether that information is indeed used during inference.
We introduce a method for analyzing how a Transformer model processes these inputs by focusing on simple arithmetic problems and their intermediate values.
To trace where information about intermediate values is encoded, we measure the correlation between intermediate values and the activations of the model using principal component analysis (PCA).
Then, we perform a causal intervention by manipulating model weights. 
This intervention shows that the weights identified via tracing are not merely correlated with intermediate values, but causally related to model predictions.
Our findings show that the model has a locality to certain intermediate values, and this is useful for enhancing the interpretability of the models.
Our code is available at {\url{www.github.com/cl-tohoku/trace-manipulate}}
\end{abstract}
\section{Introduction}

Recent language models (LMs) can solve complex input such as math word problems~\cite{saxton2018analysing,geva-etal-2020-injecting}.
To obtain the correct output from such complex (latent structured) inputs, it is necessary for multiple steps of inference via intermediate values.
However, how LMs process their inputs and capture latent structure is still not well understood.
In previous studies, \citet{linzen-etal-2016-assessing} and \citet{tran-etal-2018-importance} showed that the neural models can capture some implicit hierarchical structure, but it is unclear where that information is encoded.
\citet{shibata-etal-2020-lstm} observed that in LMs trained with Dyck language and showed some activations are highly correlated with the depth of their syntactic tree.
However, even if such features can be extracted, there is no guarantee that it is used by the model~\cite{elazar-etal-2021-amnesic,lovering2021predicting}.
Given these considerations, to better understand LM predictions for latent structured inputs, it is necessary to:
(a) To find where information about intermediate values of the latent structured inputs is encoded.
(b) To evaluate the impact of the features when the model makes predictions.

\begin{figure*}[t]
    \centering
    \includegraphics[width=\hsize]{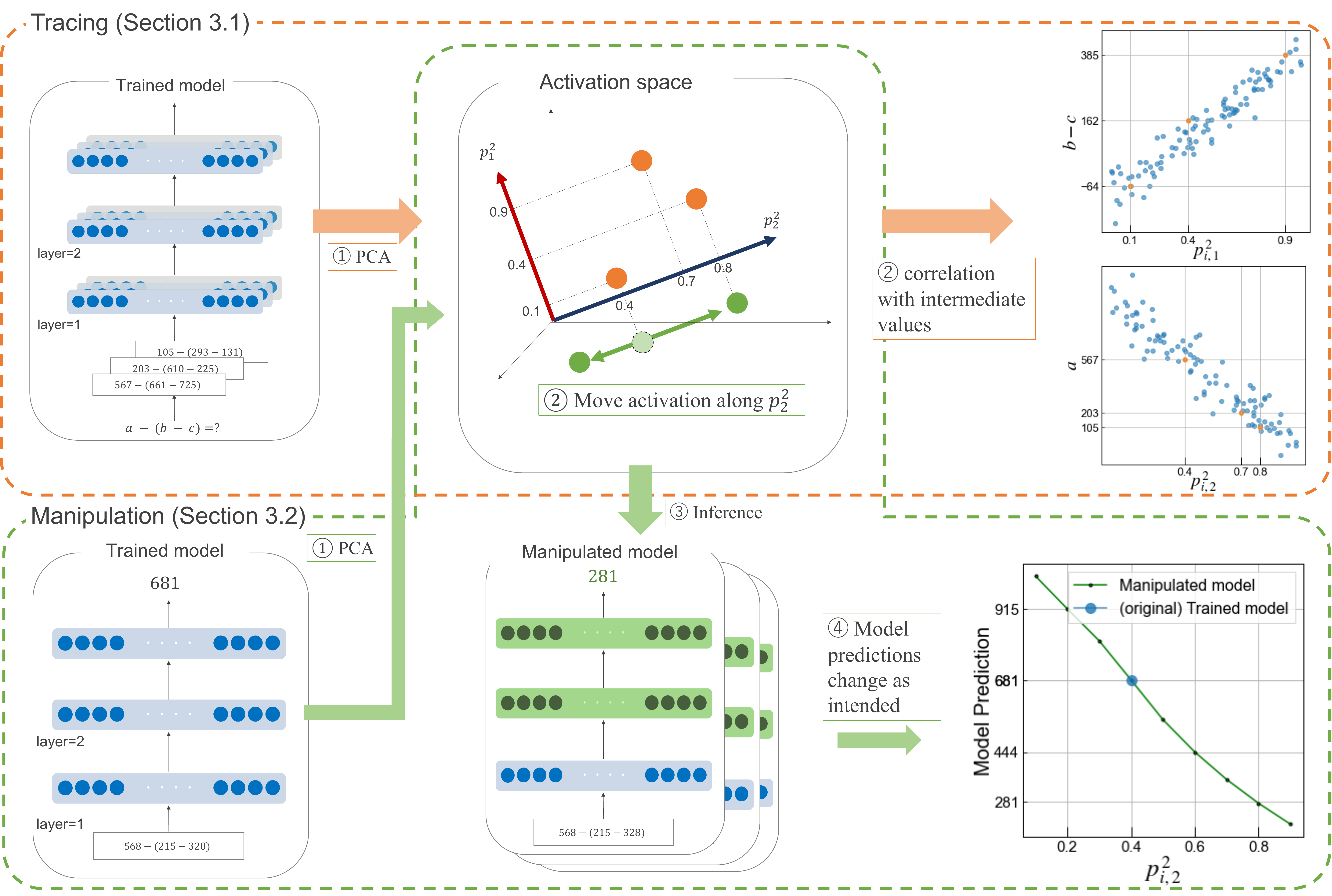}
    \caption{An overview of our methods. We find which directions obtained by PCA are correlated with intermediate values of the equations and how the model prediction changes when the weights of their directions are manipulated.}
    \label{fig:overview}
\end{figure*}
In this work, we introduce a method for analyzing the relationship between internal representations in Transformer~\cite{NIPS2017_3f5ee243}-based models and intermediate values of latent structured inputs by using simple math problems.
We choose them as a formal language because their intermediate values of the latent (tree) structure are clear and continuous, and it is easy to investigate their relationship to the internal representation of the model.
The intermediate value of $(154-38)-(290-67)$ can be clearly defined as 154, 290, $154-38=$116, and so on.
we take up a Transformer model trained to solve math equations.
An overview of our experiments is shown in Fig.~\ref{fig:overview}.
First, we search which directions of internal representations are highly correlated with intermediate values in equations by PCA (\textbf{tracing}) to find where the information about intermediate values is encoded.
We find some directions correlate very well with the intermediate values.
Second, we observe how the model prediction changed when we manipulate the weights along its direction (\textbf{manipulation}) to conduct a causal intervention.
The result of this experiment suggests that some directions of them are indeed used by the model.

These two results show that a Transformer model has a locality to certain intermediate values, and it could help enhance the interpretability of the models.
Our contributions are as follows: (a)We show that intermediate values of equations are encoded in particular directions in internal representation. (b)We show that some features representing intermediate values are used during inference.
     
    
\begin{figure}[t]
    \centering
    \includegraphics[width=\hsize]{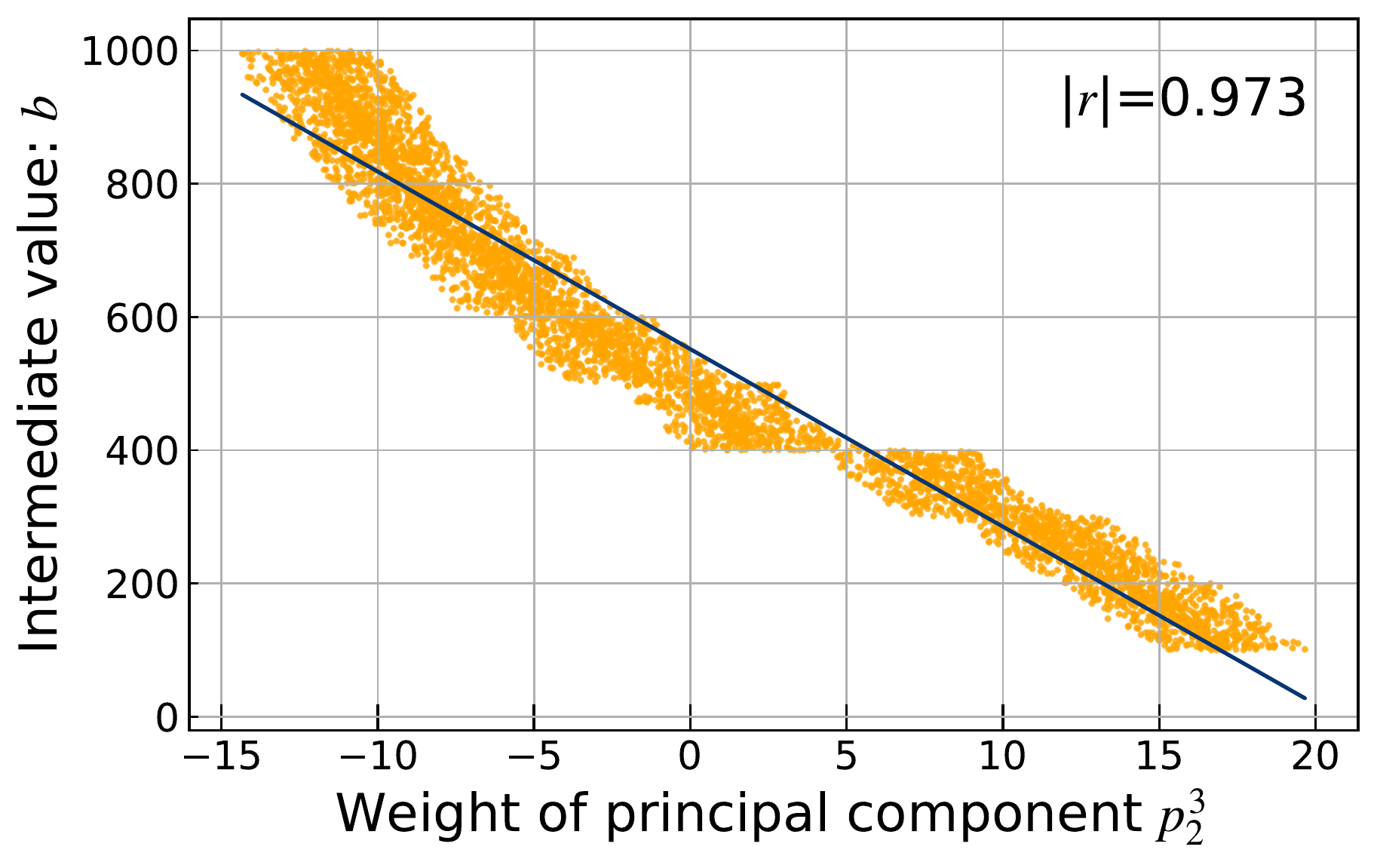}
    \vspace{-5ex}
    \caption{The relationship between $p^{3}_{i,2}$ and $R^{j}_{i}=b_{i}$ in the equation $a-(b-c)$. The correlation is very high.}
    \label{fig:pca_example}
\end{figure}
\begin{figure*}[t]
    \centering
    \includegraphics[width=\hsize]{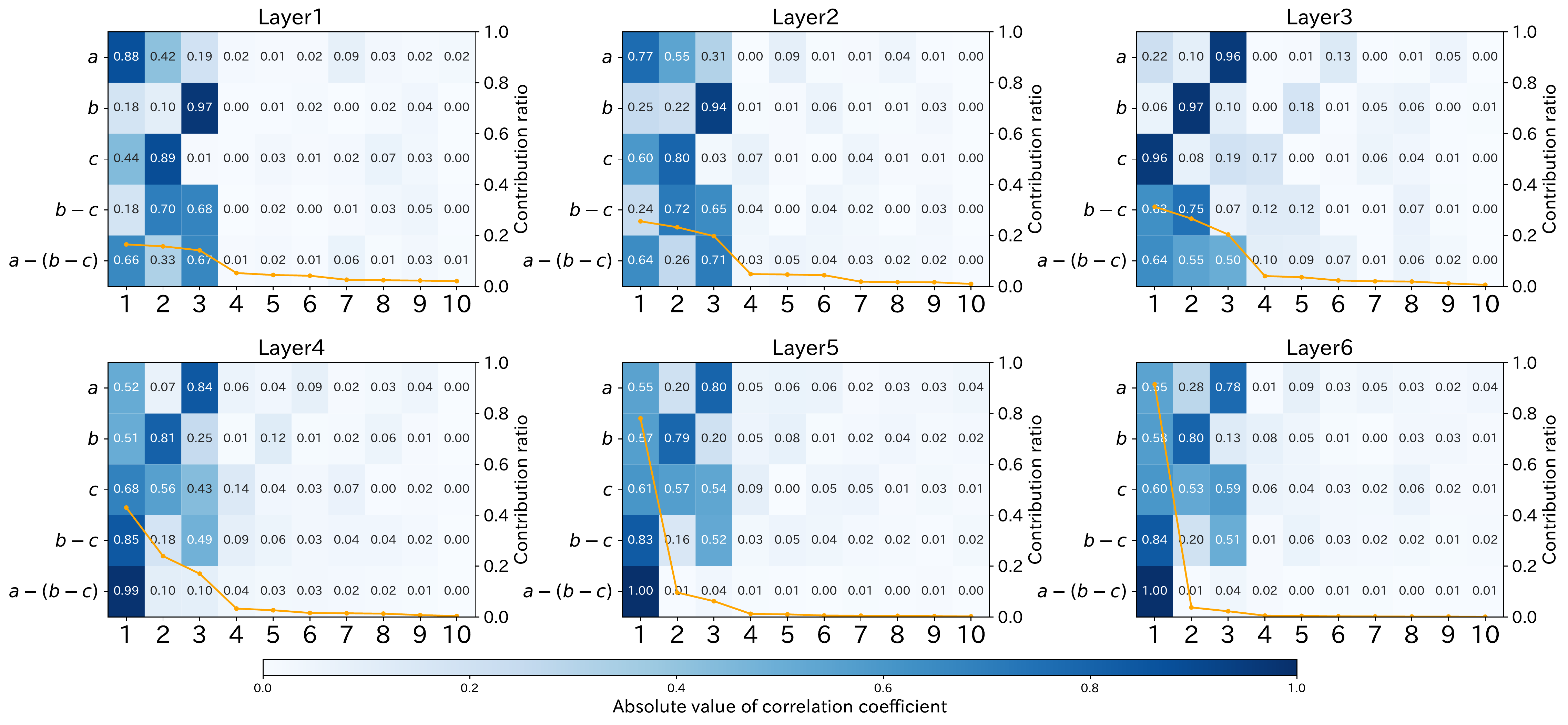}
    \caption{Correlations between each principal component and intermediate values, for all layers. Each cell represents the absolute value of the correlation coefficient between the weights of $k$-th principal component (column) and the intermediate values (row). The orange line shows the contribution ratio of each principal component.}
    \label{fig:pca_heatmap}
\end{figure*}
\section{Related Work}
\paragraph{Intermediate values.}
Previous work has examined the representation of intermediate values in neural models.
\citet{linzen-etal-2016-assessing}, \citet{bowman2015} and \citet{tran-etal-2018-importance} found that LMs capture implicit hierarchical structures to some extent, e.g., when performing logical inference over formal languages.
Closest to this work are \citet{shibata-etal-2020-lstm}, who trained LMs on the Dyck language and observed hidden units that are highly correlated with nesting depth.
In contrast to their work, we analyze representations of more complex inputs, i.e., equations, and also manipulate these representations to understand the impact of correlated activations on model predictions.
\paragraph{Numeracy}
\citet{geva-etal-2020-injecting} have shown that they can reach the state-of-the-art performance of numerical reasoning by using large pre-trained LM.
Several studies have shown that a Transformer model can solve more complex problems such as linear algebra and elementary mathematics to some extent~\citep{Francois:2021,saxton2018analysing}.
Based on their findings, we use simple mathematical equations as problems that can be solved by a Transformer model in this study.

\section{Experiments}
\begin{figure*}
\begin{subfigure}{.5\textwidth}
  \centering
  \includegraphics[width=.8\linewidth]{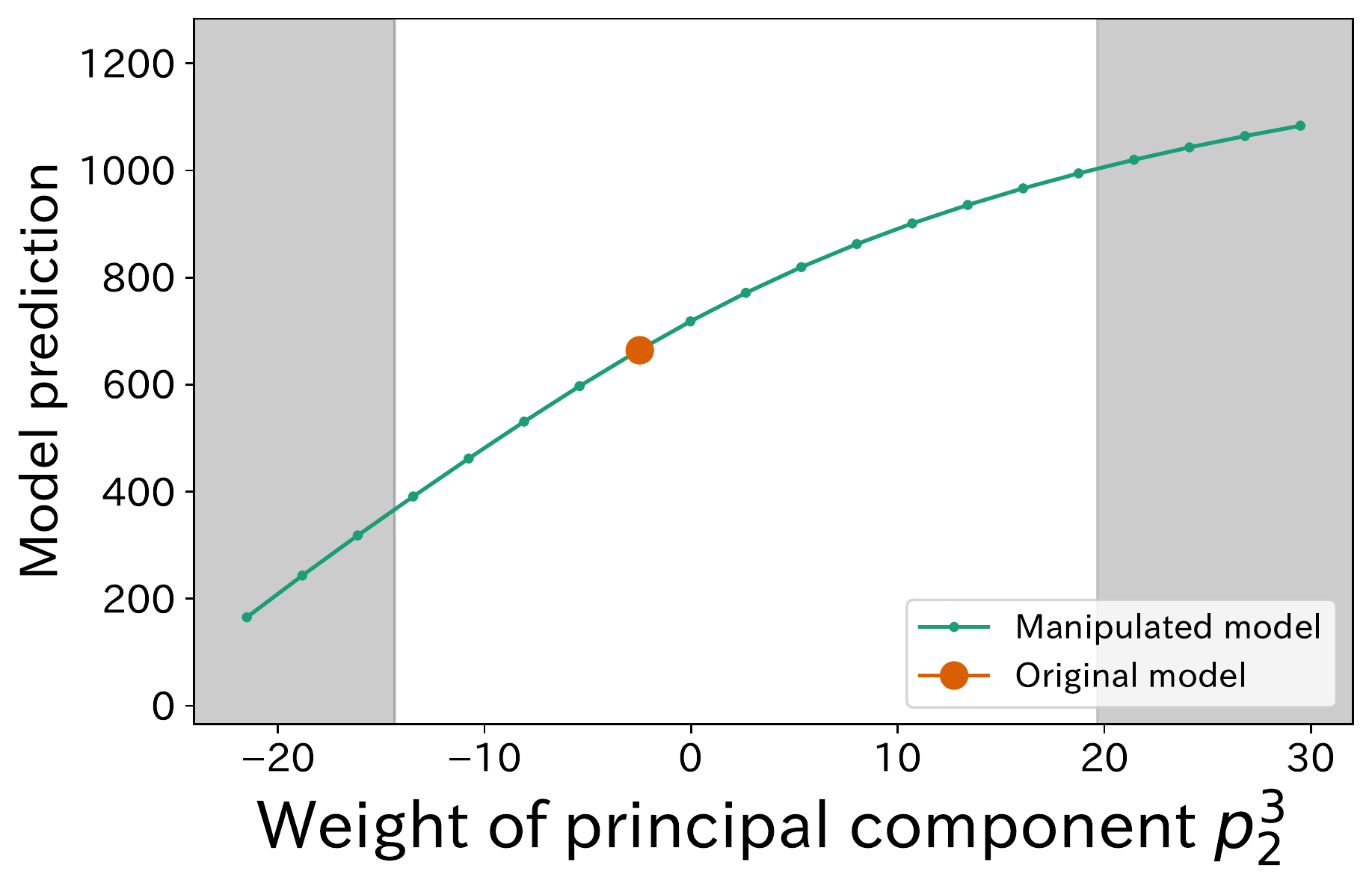}  
  \caption{Changes of model predictions as a function of weight of $p^{3}_{2}$. }
  \label{fig:manipulate_result_change}
\end{subfigure}
\begin{subfigure}{.5\textwidth}
  \centering
  \includegraphics[width=.8\linewidth]{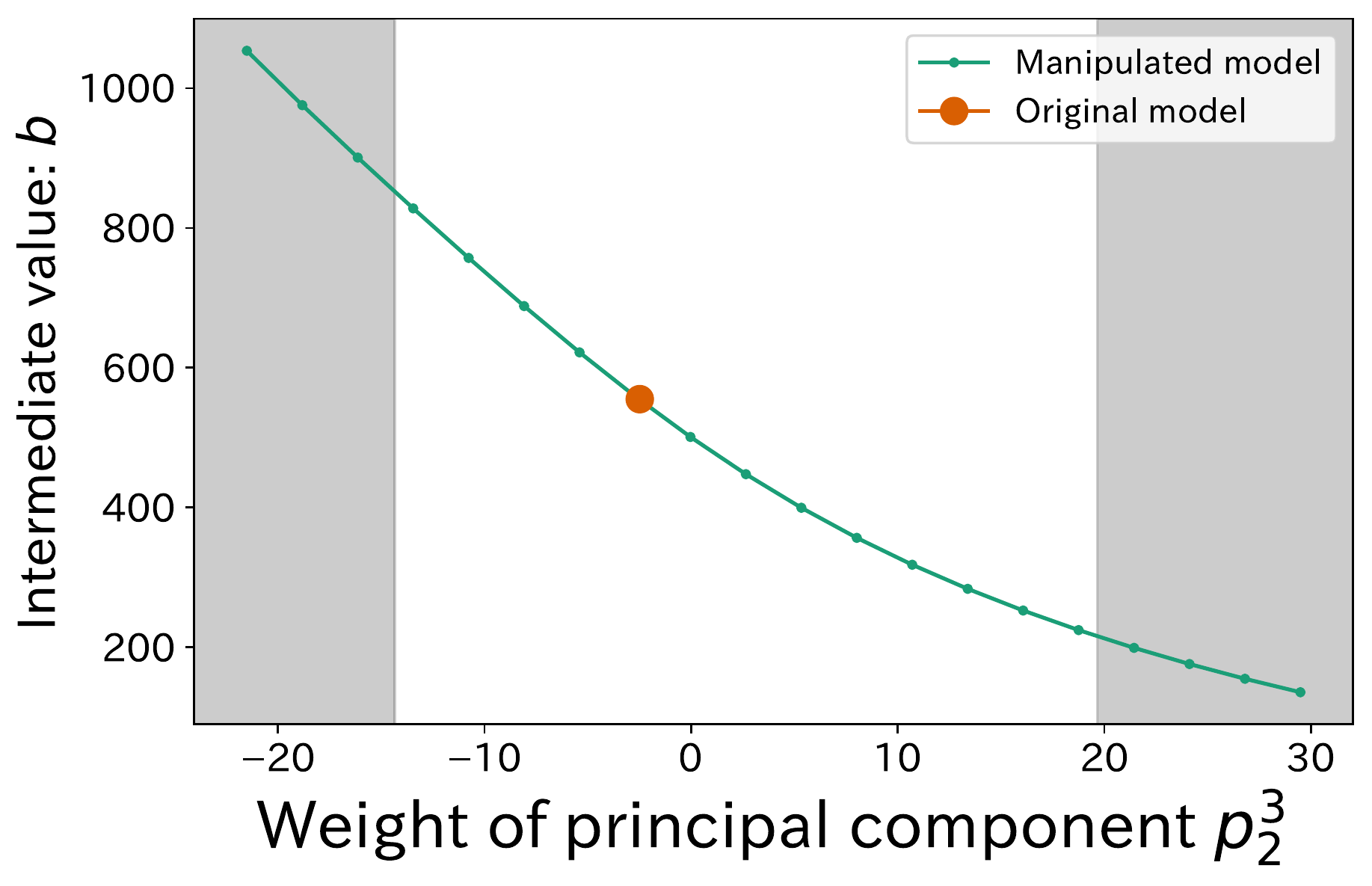}  
  \caption{The intermediate value $b$ as a function of weight of $p^{3}_{2}$.}
  \label{fig:manipulate_intermed_change}
\end{subfigure}
\newline
\begin{subfigure}{.5\textwidth}
  \centering
  \includegraphics[width=.8\linewidth]{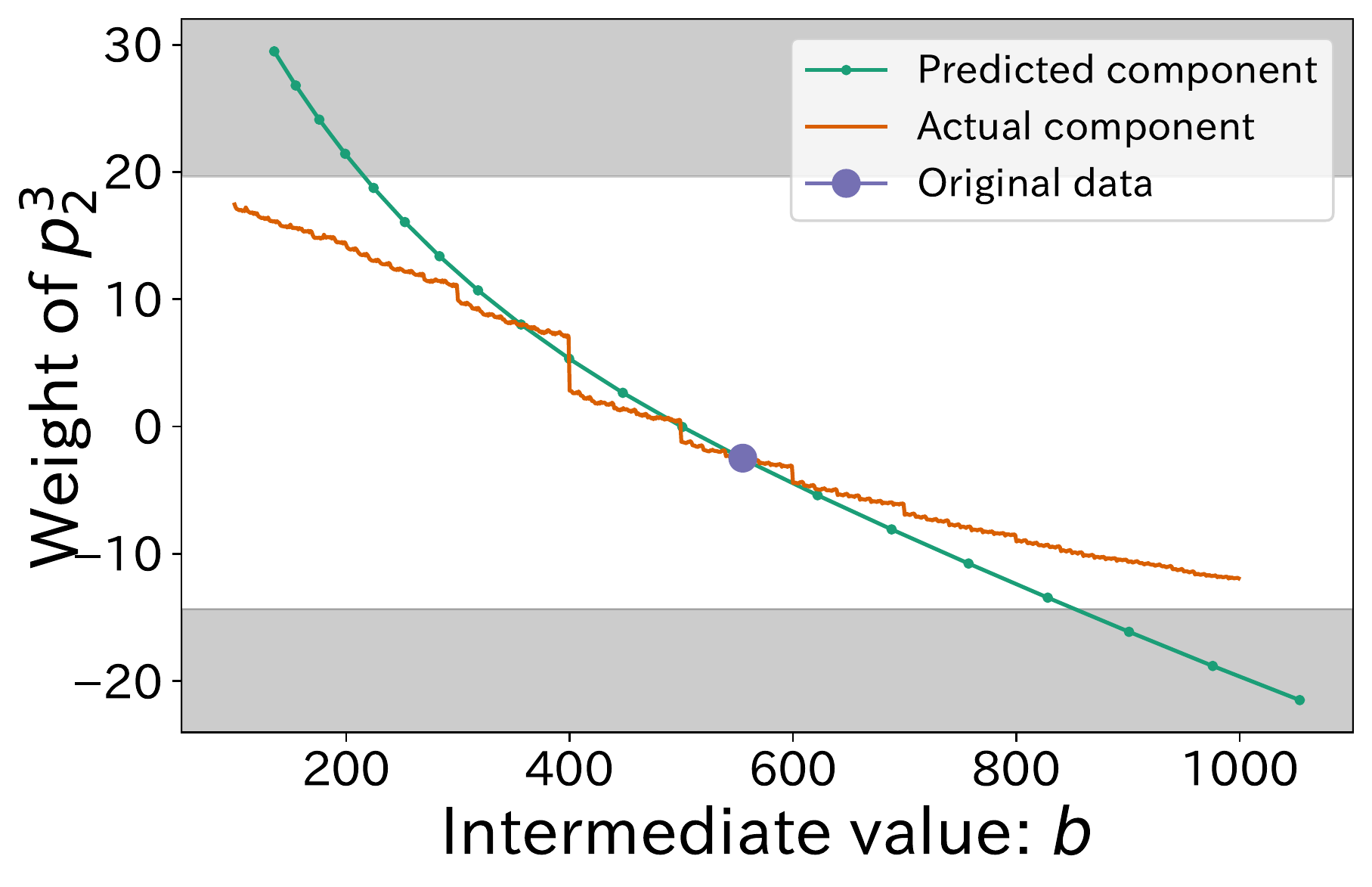}  
  \caption{Predicted and actual weights of most-correlated direction $\hat p^{3}_{2}(b)$ as a function of the intermediate value $b$.}
  \label{fig:compare_predict_and_actual_component}
\end{subfigure}
\begin{subfigure}{.5\textwidth}
  \centering
  \includegraphics[width=.8\linewidth]{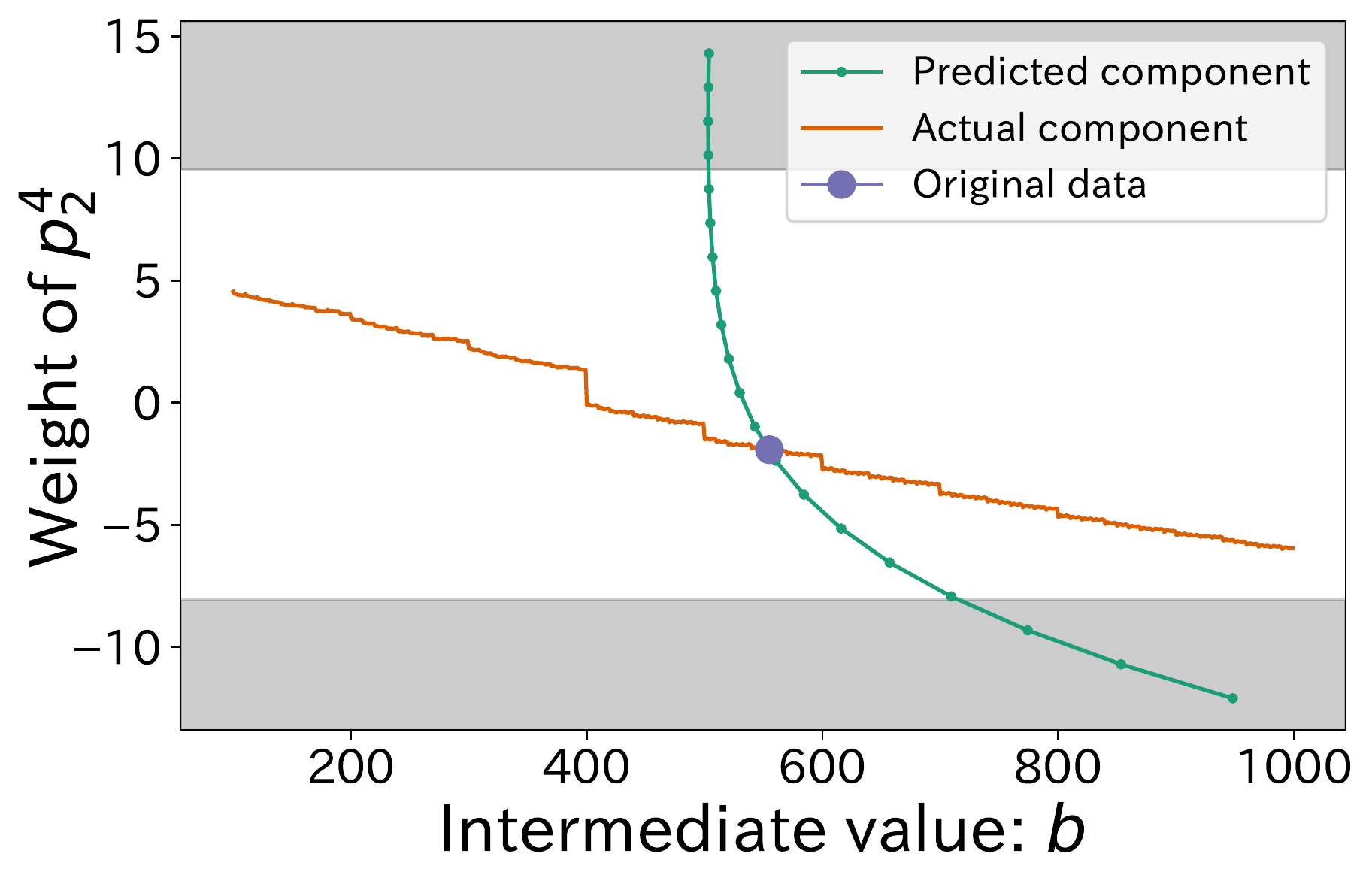}  
  \caption{Predicted and actual weights of most-correlated direction $\hat p^4_2(b)$ as a function of the intermediate value $b$.}
  \label{fig:compare_predict_and_actual_component_nouse}
\end{subfigure}
\caption{The results of manipulation. The weights of the shaded areas do not appear in the dataset. }
\label{fig:manipulation}
\end{figure*}

We conduct two types of experiments.
First we trace the representation of intermediate values in the model.
As a result we find directions in activation space that are highly correlation with intermediate values.
Then we manipulate activations along these directions and observe if model predictions change as expected.

As neural math problem solving model, we train a 6-layer Transformer using the settings by \citet{sajjad2021effect} on synthetic data.
We generate 200k equations involving up to five steps of addition or subtraction of integers between 1 and 1000, e.g., $(154-38)-(290-67)$. 
Following ~\citet{geva-etal-2020-injecting} inputs are split into digits, e.g., ``$123$'' is tokenized into $1 ,\#\#2 ,\#\#3$.
Model predictions are obtained via linear regression on the final layer's [CLS] token representation.
After training on 190k equations we evaluate the model on 10k equations and obtain a regression score of $R^2 = 0.9988$, i.e., the model solves the equations almost perfectly.

\subsection{Tracing intermediate values}

\label{subsec:tracing_experiment}
\paragraph{Method.}
We describe our method for tracing the representation of intermediate values in model activations.
First, we reduce the dimensionality of the activations at each model layer.
Let $h^{l}_{j}$ be the layer activations of the $j$-th word in a hidden layer $l$. 
Given an input of length $n$, we concatenate all token representations in layer $l$, obtaining the layer representation $H^l = h^{l}_{1}\oplus h^{l}_{2}\oplus \cdots \oplus h^{l}_{n}$ and fit a PCA to obtain the top 10 principal components $p^{l}_{k},k \in [1, ..., 10]$. Applying this PCA to instance $i$ yields the 10-dimensional representation $ p^{l}_{i,k}$.
Our hypothesis is that the intermediate values are encoded by one or more of the principal components.
Intuitively, we assume that a principle component encodes an intermediate value if the magnitude of model activation in this direction correlates with the magnitude of the intermediate values.
To test this hypothesis, we measure the correlation $\mathbf{corr}(R^{j}_i, p^{l}_{i,k})$ between the value of the intermediate values $R^{j}_{i}$ and the magnitude of principal component $k$ in the representation $p^{l}_{i,k}$.
Finally, we obtain \textbf{most-correlated direction} $\hat{p}^l_k(R^{j}) := \mathbf{argmax}_k(\mathbf{corr}(R^{j}_i, p^{l}_{i,k}))$.
If this correlation is high, we conclude that the intermediate value is encoded in that direction.
\paragraph{Results.}
We trace intermediate values for the equation pattern $a-(b-c)$.
For example, Fig.~\ref{fig:pca_example} shows a strong correlation of $0.973$ between the intermediate value $b$ and its most-correlated direction $p^{3}_{2}$.
After measuring the correlation of each intermediate value and each of the top 10 principal components, we plot all correlations in Fig.~\ref{fig:pca_heatmap}.
Overall, most-correlated directions show high correlations with intermediate values with moderate contribution ratio up to the 3rd layer, which we take as evidence that the model encodes intermediate values along these directions.

\subsection{Manipulating intermediate values}
\paragraph{Method.}
\label{subsec:manipulating_experiment}
So far, we found correlations between intermediate values and directions in activation space.
However, such correlations do not necessarily mean that these directions determine model predictions.
To test if the directions we found actually influence model predictions, we perform causal interventions by \emph{manipulating} activations.
Concretely, we manipulate activations along principal components and observe changes in model predictions, as shown in Fig.~\ref{fig:overview}.
Formally, we transform layer representation $H^{l}$ (see \S\ref{subsec:tracing_experiment}) into $H^{l \prime}$, by increasing or decreasing its projection onto the principal component $p^l_k$ by a factor of $r$:
\begin{align}
 H^{l \prime } \leftarrow H^l + (r-1)\left({p^{l}_k}^{\top} H^l\right) p^l_k
\end{align}
Intuitively, increasing $r$ moves $H^{l}$ along $p^l_k$.

If a most-correlated direction $\hat{p}^l_{k}(R^j)$ indeed encodes the intermediate value $R^j$, it should be possible to manipulate activations in a way that corresponds to changing $R^j$.
For example, if the model prediction given the input $43-(50-20)$ changes from the 13 to 19, this difference is consistent with changing the first input term from 43 to 49.
By manipulation factors $r$ of a particular most-correlated direction, observing model predictions, and calculating corresponding intermediate values, we obtain data for fitting a function from intermediate values to manipulation factors $r$. That is, we learn to manipulate activations in a way that corresponds to changing a particular intermediate value.
To assess the fidelity of this manipulation, we change input terms and compare \emph{actual} activation changes along the most-correlated direction $\hat{p}^l_{k}(R^j)$ to the factor $r$ \emph{predicted} by our fitted function.


\paragraph{Results.}
Using the input $617-(555-602)$ and the intermediate value $b = 555$ as example, we find its most-correlated direction $\hat{p}^{3}_{2}(b)$, as described in \S\ref{subsec:tracing_experiment}.
By manipulating activations along $p^3_{2}$, model predictions change from the original 664 to results ranging from ca. 200 to 1000, as shown in Fig.~\ref{fig:manipulate_result_change}.
Calculating intermediate values $b$ that are consistent with these model predictions, we obtain Fig.~\ref{fig:manipulate_intermed_change}.
By axis inversion we obtain a function from $b$ to \emph{predicted} manipulation factors $r$ for component $p^3_{i,2}$.
We compare these \emph{predicted} component weights to the \emph{actual} component weights observed under changed inputs $\{(617-(i-602)) | (100 \leq i<1000)\}$ (Fig.~\ref{fig:compare_predict_and_actual_component}).
Predicted and the actual weights of the most-correlated direction agree well (corr. 0.986, $R^2$ score 0.687), which we take as evidence that $\hat{p}^{3}_{2}(b)$ encodes the intermediate value $b$ and determines model predictions accordingly.
Conversely, manipulation identifies most-correlated directions that are correlated but less used in prediction.
The most-correlated direction $\hat{p}^{4}_{2}(b)$ has a high correlation of 0.81 with $b$, but predicted component weights show much less agreement with actual weights (corr.\ 0.802, $R^2$ score $-1.06\times 10^4$, Fig.~\ref{fig:compare_predict_and_actual_component_nouse}).

In conclusion, this case study showed how manipulations in activation space can find a causal connection to intermediate values.

\section{Acknowledgments}
This work was supported by JST CREST Grant Number JPMJCR20D2, Japan.
\bibliography{anthology}




\end{document}